\documentclass[preprint,12pt]{elsarticle}

\usepackage[T1]{fontenc}
\usepackage{amsmath,amssymb,amsfonts}
\usepackage{algorithmic}
\usepackage{graphicx}
\usepackage{textcomp}
\usepackage{xcolor}
 \usepackage{caption}
\usepackage{subcaption}
\usepackage{cleveref}
\usepackage{booktabs}
\usepackage{float}

\usepackage{amssymb}
\usepackage{amsmath}
\usepackage{tikz}

\def\layersep{3cm}
\newcommand\nn[1]{
    \foreach \y in {1,2,3,5}
        \node[neuron, fill=green!40] (X\y-#1) at (0 ,\y+1) {$X\y$};

    \foreach \y in {1,...,5}
        \path node[neuron, fill=blue!40] (H\y-#1) at (\layersep,\y) {$H\y$};

    \node[neuron, fill=red!40] (Y-#1) at (2*\layersep,2.5) {$Y$};

    \foreach \source in {1,2,3,5}
        \foreach \dest in {1,...,5}
            \path (X\source-#1) edge (H\dest-#1);

    \foreach \source in {1,...,5}
        \path (H\source-#1) edge (Y-#1);
}

\journal{}

\begin{document}

\begin{frontmatter}



\title{Deep of Recurrent Neural Networks and Bayesian Neural Networks for Detecting Electric Motor Damage Through Sound Signal Analysis}


\author{Waldemar Bauer$^a$, Jerzy Baranowski$^a$} %% Author name

\affiliation{organization={Dept. of Automatic Control \& Robotics},%Department and Organization
            addressline={AGH University of Krakow}, 
            %city={Krakow},
            country={Poland}}

%% Abstract
\begin{abstract}
Fault detection in electric motors is a critical challenge in various industries, where failures can result in significant operational disruptions. This study investigates the use of Recurrent Neural Networks (RNNs) and Bayesian Neural Networks (BNNs) for diagnosing motor damage using acoustic signal analysis. A novel approach is proposed, leveraging frequency domain representation of sound signals for enhanced diagnostic accuracy. The architectures of both RNNs and BNNs are designed and evaluated on real-world acoustic data collected from household appliances using smartphones. Experimental results demonstrate that BNNs provide superior fault detection performance, particularly for imbalanced datasets, offering more robust and interpretable predictions compared to traditional methods. The findings suggest that BNNs, with their ability to incorporate uncertainty, are well-suited for industrial diagnostic applications. Further analysis and benchmarks are suggested to explore resource efficiency and classification capabilities of these architectures.
\end{abstract}


% \begin{highlights}
% \end{highlights}

%% Keywords

\end{frontmatter}



\section{Introduction}
Detecting faults is a critical task in various industrial sectors, including manufacturing, transportation, and energy production. Failures in these systems can result in severe consequences such as operational breakdowns, higher maintenance expenses, and even threats to human safety. Despite significant research efforts in recent years, creating cost-effective and dependable classifiers for fault detection in electric motors remains a challenge. One major obstacle is gathering a well-balanced dataset that includes signals from both operational and faulty equipment. To address this issue, the use of a Bayesian neural network (BNN) is suggested, as it is particularly effective in handling imbalanced training data.

Currently, the detection of motor faults is predominantly addressed through deep neural networks (DNNs) utilizing various signal types. The study by \cite{Tran2023} demonstrates the application of DNNs for fault detection in servo mechanism systems based on acceleration measurements. Similarly, the authors of \cite{Qiu2023} propose a comparable approach, leveraging current signal analysis for fault identification. Additionally, vibration signals have been explored as a viable input for fault detection, as exemplified in \cite{Tama2022}. An alternative methodology incorporating data fusion techniques for fault diagnostics is presented in \cite{Zhang2023}. Furthermore, \cite{Zhu2023} provides a comprehensive review of machine learning-based approaches for fault detection in electric motors, offering insights into the advancements and challenges within this domain.

Bayesian neural networks (BNNs) are a class of neural networks that incorporate Bayesian inference to estimate the probabilistic distribution of network weights \cite{mac2003}. By integrating prior knowledge of the data and propagating uncertainty throughout the network, BNNs enhance predictive robustness and provide a more reliable quantification of uncertainty in the input data \cite{10.5555/3045390.3045502, Hinton2012}. In recent years, BNNs have been increasingly utilized for fault detection across various industrial applications, with multiple studies demonstrating their efficacy. For example, \cite{Bakri2017} introduced a BNN-based fault detection framework for wind turbines capable of handling sparse data while improving uncertainty estimation. The proposed system exhibited superior performance compared to conventional methods, significantly reducing false alarms. Similarly, \cite{Loboda2015} presented a BNN-based fault detection approach for gas turbine engines, demonstrating enhanced fault detection accuracy over traditional neural networks. Furthermore, \cite{Ma2020} proposed a fault detection system for chemical processes employing BNNs with dropout, which yielded more precise fault detection outcomes than conventional neural network-based methods.

A Deep Neural Network (DNN) represents a sophisticated class of artificial neural networks characterized by multiple layers of interconnected neurons. These networks are designed to capture complex patterns and dependencies within data through hierarchical feature learning, where each successive layer extracts increasingly abstract representations. The term "deep" signifies the presence of multiple hidden layers, which enable DNNs to achieve state-of-the-art performance in various domains, including image recognition, natural language processing, and predictive analytics. By employing activation functions and optimization techniques such as backpropagation, DNNs iteratively refine their parameters to enhance predictive accuracy and generalization.

Although prior research has demonstrated the theoretical efficacy of Bayesian Neural Networks (BNNs) in fault detection, limited attention has been given to the challenges associated with signal availability and representation in this context. Addressing these limitations, the development of a robust classifier capable of performing rapid fault diagnosis using acoustic data acquired from readily accessible devices, such as smartphones, presents a promising avenue for enhancing fault detection methodologies in practical applications.

Thus, we compare a Bayesian neural network (BNN) with Deep neural network  based (DNN) on frequency domain instead of time domain is proposed in this paper. The contributions of the paper are:
\begin{itemize}
\item A method was proposed for acoustic signal representation for damage detection purposes based on frequency domain,
\item A structure of the Bayesian neural network for detecting damage was proposed and analysed,
\item A structure of the recurrent neural network  for detecting damage was proposed and analysed,
\item comparison result between proposed neural network architectures.
\end{itemize}
All proposed architectures have been validated on real sound signals collected using a smartphone from domestic appliances.
 
The paper is arranged in the following manner. \Cref{sec:BNN} provide an overview of the Bayesian inference approach used throughout the paper, covering the fundamentals of the Bayesian Neural Network method, data set characteristics and the schema of used Bayesian Neural Network. Then we follow with the description of considered case study in \Cref{sec:CaseStudy} and results of BNNs on the sound signals set from real devices \Cref{sec:FaultDetection}. We finish the paper with the conclusions in \Cref{sec:Conclusions}.

\section{Bayesian neural networks}
\label{sec:BNN}

Bayesian Neural Networks (BNNs) represent a class of neural networks that integrate Bayesian inference to enable a probabilistic learning framework. Unlike conventional neural networks, which rely on fixed point estimates for weight parameters, BNNs learn a probability distribution over these weights. This probabilistic representation facilitates the incorporation of prior knowledge and uncertainty into the model, enhancing its robustness and generalization capabilities \cite{Rezende2014, mac2003}.

Suppose we have a training dataset:
\[
D = \lbrace(x_1, y_1), (x_2, y_2), …, (x_n, y_n)\rbrace,
\]
where $x_i$ is the $i-$th input and $y_i$ is the corresponding output. Let $w$ denote the weights of the neural network. In a traditional neural network, we would try to find a set of weights $w$ that minimize the loss function $L(D,w)$ over the training dataset:

\begin{equation}
      L(D,W) = \sum_{x_i,y_i}(y_i - f(x_i,w))^2+ \lambda\sum_{d}w_{d}^2 
\end{equation}

In a BNN, we instead learn a probability distribution over the weights:

\begin{equation}
    \mathrm{log}\,p(D,w) = \sum_{x_i,y_i}\mathrm{log}N(y_i | f(x_i,w),I)\\ + \sum_{d}\mathrm{log}N(w_d |0,\cfrac{1}{\lambda})
\end{equation}
where $p(D,w)$ is the likelihood of the data given the weights, $p(w)$ is the prior distribution over the weights, and $p(D)$ is the marginal likelihood of the data. 

BNNs integrate uncertainty and prior knowledge into neural network learning by employing Bayesian inference to estimate a probability distribution over the weights. This distribution enables the model to generate predictions while quantifying uncertainty, providing insights into the confidence of the learned parameters. Using Bayes' theorem, we can compute the posterior distribution over the weights:
\begin{equation}
\begin{aligned}
p(w|D) =&{} \cfrac{p(D|w)p(w)}{p(D)}\\ =&{} \cfrac{p(D|w)p(w)}{\int_{w'}p(D|w')p(w')\mathrm{d}w'} 
\end{aligned}
\end{equation}
The posterior distribution gives us a distribution over possible values for the weights, which we then use to make predictions and calculate uncertainties.

To make predictions on a new input $x$, we compute the posterior predictive distribution over the output $\hat{y}$:
\begin{equation}
\label{eq:Integral}
\begin{aligned}
p(\hat{y}(x)|D) =&{} \int_{w} p(\hat{y}(x)|w)p(w|D)\mathrm{d}w\\ =&{}\mathbb{E}_{p(w|D)}[p(\hat{y}(x)|w]
\end{aligned}
\end{equation}
The integral in \eqref{eq:Integral} averages the predictions over all possible values of the weights, weighted by their posterior probabilities.

\section{Case study} 
\label{sec:CaseStudy}

\subsection{Dataset and faults}

\begin{figure}[htb]
     \centering
          \begin{subfigure}[b]{0.42\textwidth}
         \centering
  \includegraphics[width=1\columnwidth]{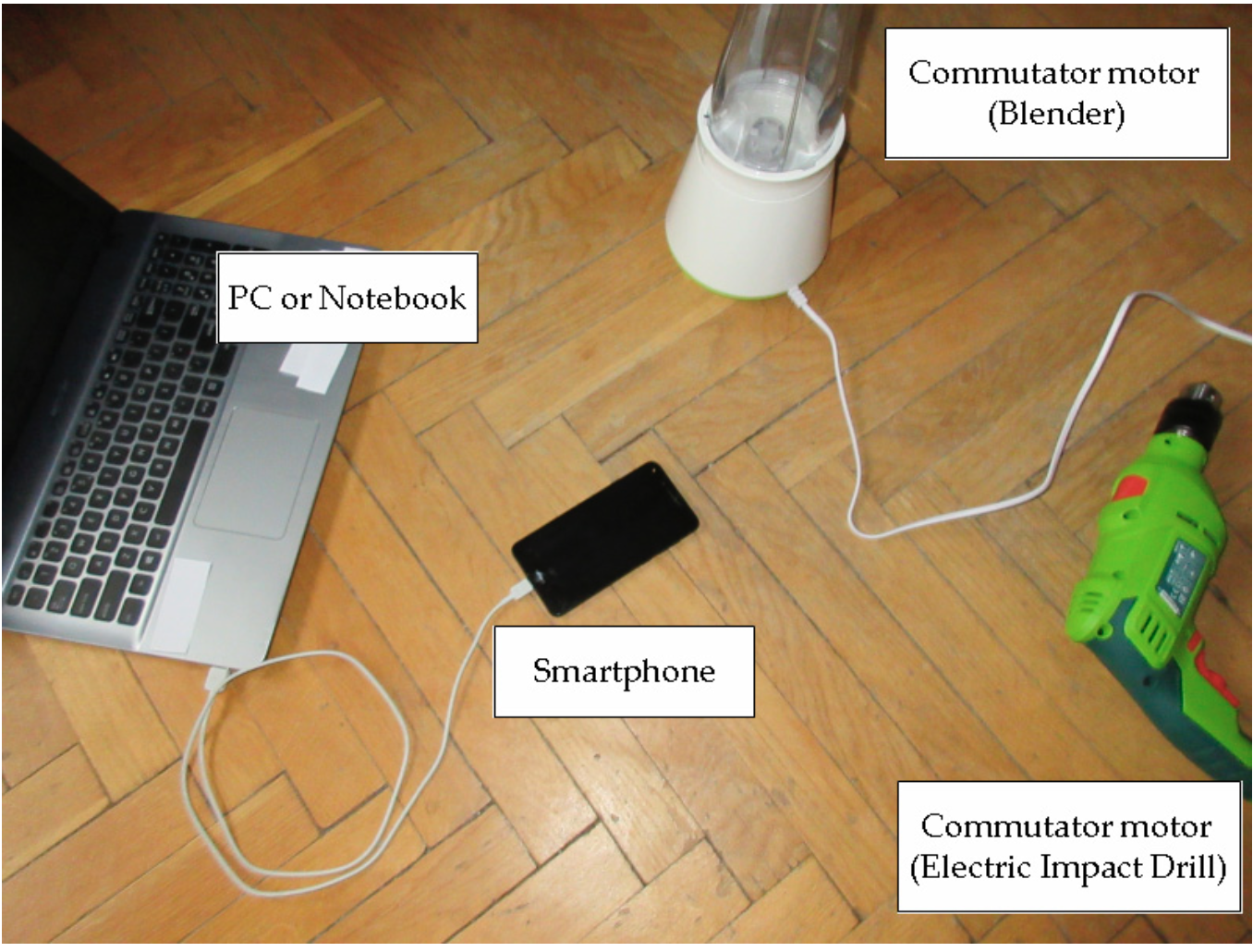}
\caption{Experimental setup\label{fig:aqui}}
     \end{subfigure}
     \begin{subfigure}[b]{0.42\textwidth}
     \centering
  \includegraphics[width=1\columnwidth]{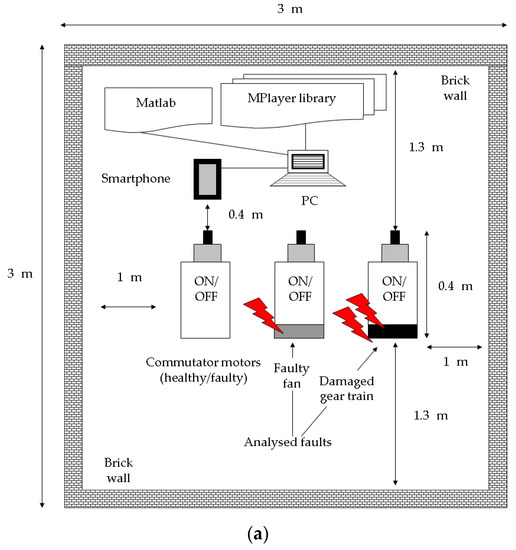}
\caption{Data acquisition setup\label{fig:aqui_1}}
     \end{subfigure}
     \hfill
        \caption{Acoustic data acquisition from blender and electric impact drill using a smartphone (CC BY 4.0, source \cite{glowacz2018})}
        \label{fig:DataAcquisition}
\end{figure}

\begin{figure*}[!htb]
     \centering
     \begin{subfigure}[!tb]{0.42\textwidth}
         \centering
  \includegraphics[width=1\columnwidth]{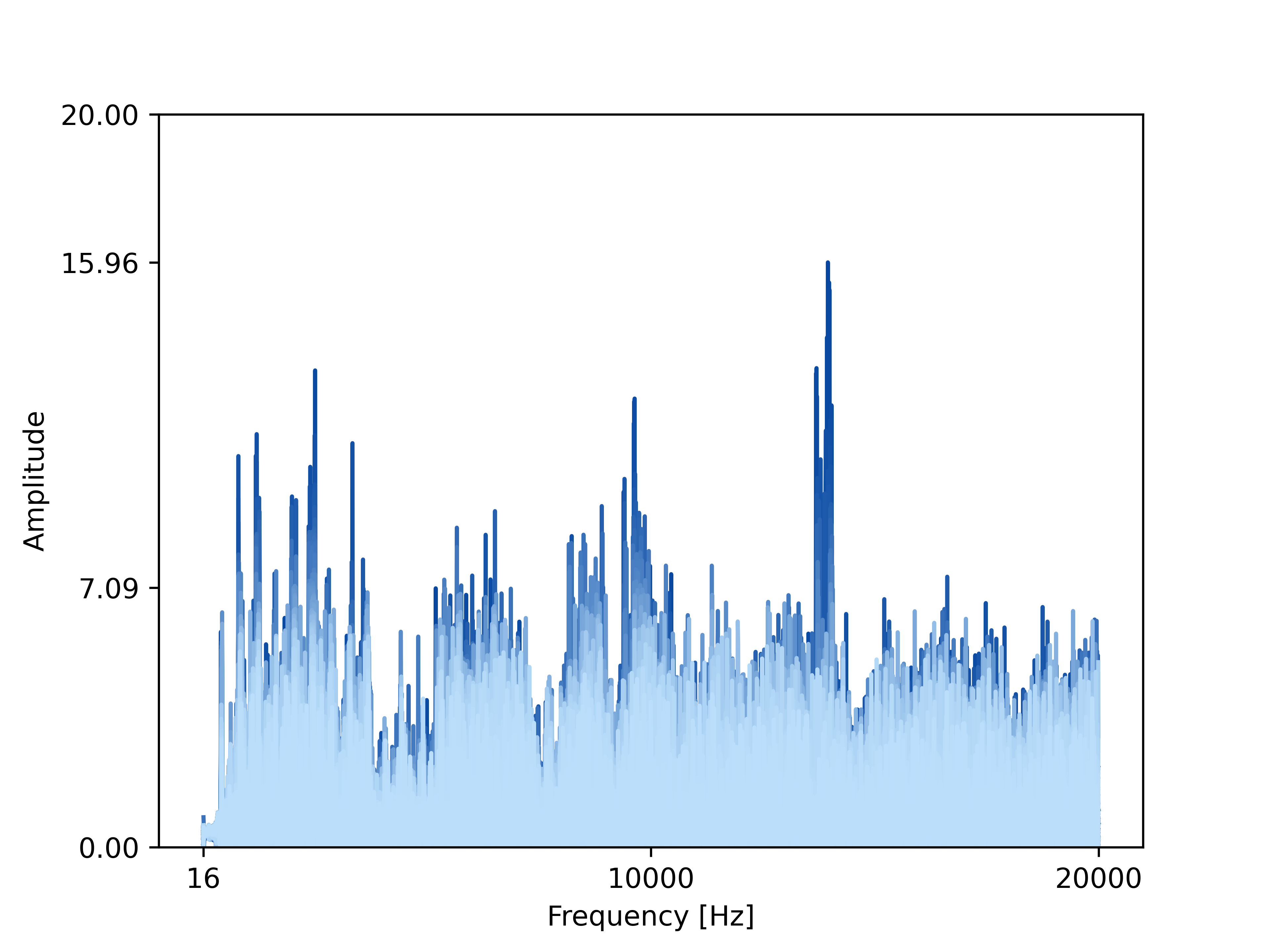}
\caption{FFT transforms for signals of properly functioning devices.\label{fig:hsig}}
     \end{subfigure}
     \hfill
     \begin{subfigure}[!tb]{0.42\textwidth}
         \centering
  \includegraphics[width=1\columnwidth]{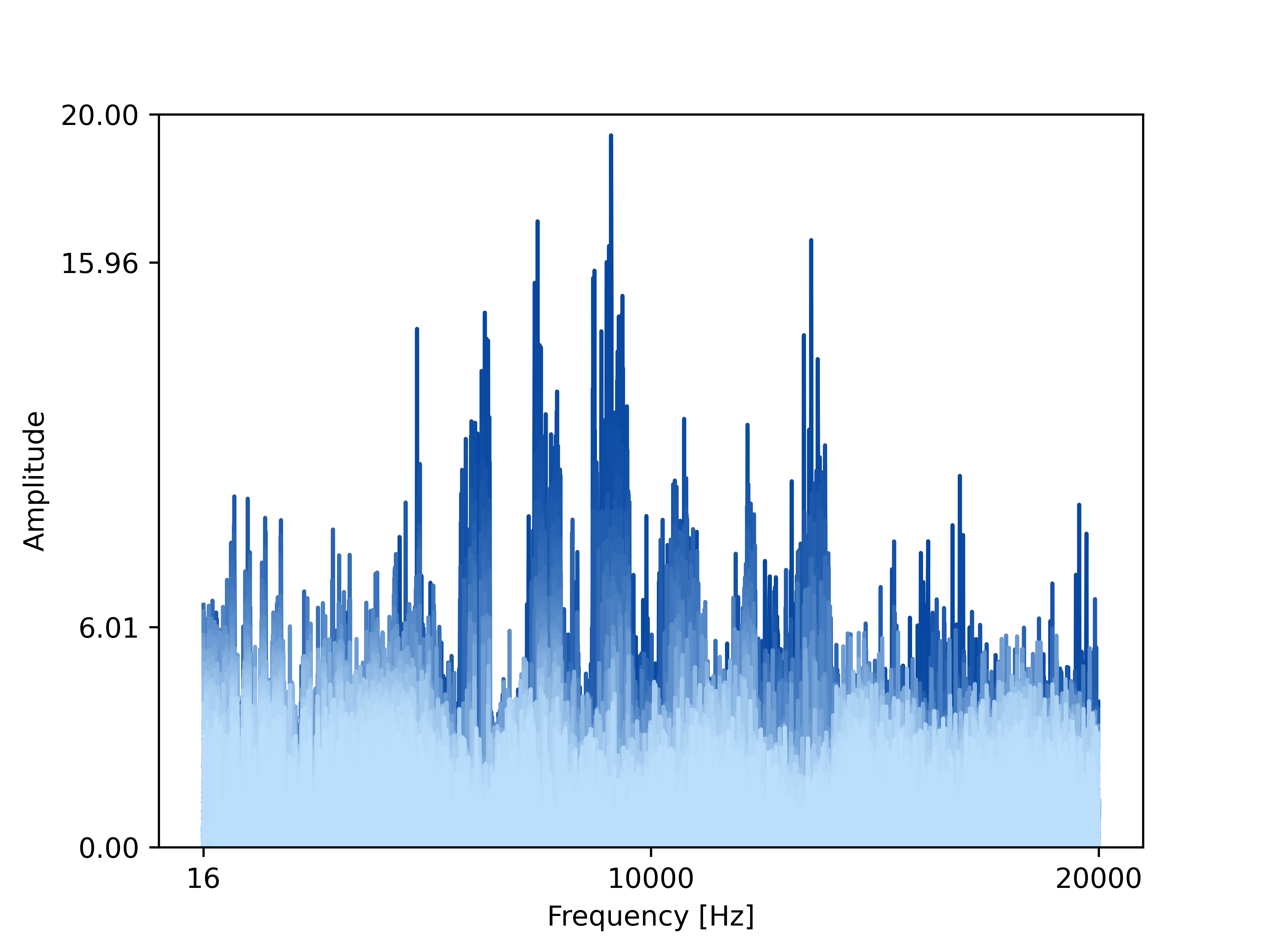}
\caption{FFT transforms for signals of faulty devices. \label{fig:uhsig}}
     \end{subfigure}
     \hfill
        \caption{FFT transforms. The light blue color of the plot represents signals with small maximum amplitudes, while the dark blue color corresponds to signals with large maximum amplitudes}
        \label{fig:FFT}
\end{figure*}

The experiments will utilize a dataset obtained from the publication by \cite{glowacz2018}. This dataset comprises recordings that have been categorised into two distinct classes: 
\begin{enumerate}
    \item Healthy device, 
    \item Device with a damaged motor, with subcategories:
    \begin{itemize}
        \item Device with damaged gear train (Fault 1),
        \item Device with 5 broken rotor blades (Fault 2),
        \item Device with 10 broken rotor blades (Fault 3), 
        \item Device with a shifted brush (motor off) (Fault 4).
        \end{itemize}
\end{enumerate}
Each class contains 30 one-second-long recordings, which are saved as \texttt{.wav} files with a sampling rate of 4400 Hz. The undamaged device set for the experiment is selected as the first set, while the other sets are used to describe the device with damages. The acquisition of acoustic data from blender and electric impact drill is illustrated in Figures \ref{fig:aqui_1} and \ref{fig:aqui}.

% \begin{figure}[!t]
% \centering
%   \includegraphics[width=1\columnwidth]{diag_accu.jpeg}
% \caption{Schema of acoustic data acquisition using smartphone (CC BY 4.0, source \cite{glowacz2018})\label{fig:aqui_1}}
% \end{figure}

% \begin{figure}[!t]
% \centering
%   \includegraphics[width=1\columnwidth]{aqu.png}
% \caption{Experimental setup to acoustic data acquisition from blender and electric impact drill (CC BY 4.0, source \cite{glowacz2018})\label{fig:aqui}}
% \end{figure}

\subsection{Frequency domain analysis}
To perform the analyses, the transition to the frequency domain was chosen due to the variable lengths of time-series data, such as audio signals. Additionally, the high sampling rate results in a large number of samples within even short segments, making time-domain analysis more complex. Analyzing signals in the frequency domain also eliminates issues related to phase shifts.

For the experimental setup, each signal was segmented into equal one-second intervals, ensuring sufficient information retention in the Fast Fourier Transform (FFT). The FFT was computed for each signal, with the frequency range constrained to the audible spectrum (16 Hz–20 kHz) to align with the microphone's characteristics used in the case study \cite{porkeba2022functional}.

The FFT transformations (Fig. \ref{fig:FFT}) corresponding to signals from properly functioning devices are illustrated in Fig. \ref{fig:hsig}, while Fig. \ref{fig:uhsig} displays FFT results for faulty devices. The key differences between these cases are primarily observed in the maximum amplitude and the overall shape of the spectral characteristics, as shown in Fig. \ref{fig:FFT}.

\section{Fault detection results}
\label{sec:FaultDetection}
In this section we will compare the structure and correctness of the developed neural networks.

\subsection{Deep neural network}
This network has 2025 inputs to which subsequent samples of the calculated FFT from the classified signal are fed. Then the signal is fed to hidden layers which successively narrow the size of the analyzed data in the following order 180,60, 60, 30. Of course, the connections between neurons have weights learned in the learning process and an all-to-all connection was used between neurons. The relu function was used as the activation function between layers. ConnectionsOn the other hand, the network has one output whose activation function is the sigmoid function.

\begin{figure}
    \centering
    \includegraphics[width=0.7\linewidth]{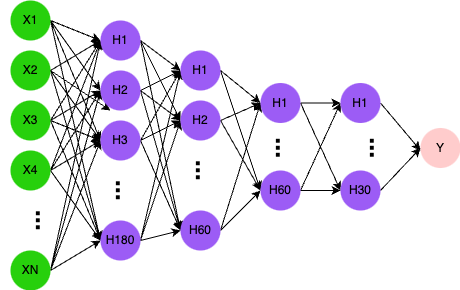}
    \caption{Structure of the Deep neural network used in the
experiments. The developed network takes the successive values of the calculated FFT signal as input arguments, X1,...,XN, which are passed to fore hidden layers, whose activation function is relu. The output of the network, Y, uses the sigmoid functionas an activation function.}
    \label{fig:dnn}
\end{figure}

% \begin{figure}[!b]
% \centering
%   \includegraphics[width=1\columnwidth]{hsig_fft.png}
% \caption{FFT transforms for the set containing signals of properly functioning devices. The light blue color of the plot represents signals with small maximum amplitudes, while the dark blue color corresponds to signals with large maximum amplitudes.\label{fig:hsig}}
% \end{figure}

% \begin{figure}[!t]
% \centering
%   \includegraphics[width=1\columnwidth]{uhsig_fft.png}
% \caption{FFT transforms for the set containing signals of properly functioning devices. The light blue color of the plot represents signals with small maximum amplitudes, while the dark blue color corresponds to signals with large maximum amplitudes.\label{fig:uhsig}}
% \end{figure}

\subsection{Structure of the Bayesian neural network}
\label{str_bnn}
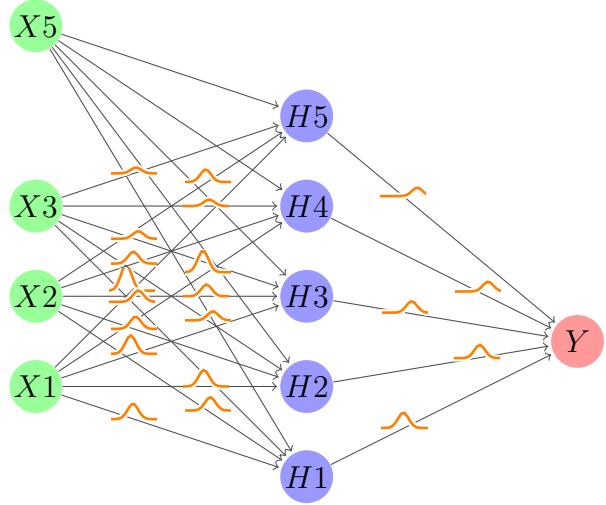
\begin{figure}[!hb]
    \centering
    
\begin{tikzpicture}[
    scale=1.2,
    shorten >=1pt,->,draw=black!70, node distance=\layersep,
    neuron/.style={circle,fill=black!25,minimum size=20,inner sep=0},
    edge/.style 2 args={pos={(mod(#1+#2,2)+1)*0.33}, font=\tiny},
    distro/.style 2 args={
        edge={#1}{#2}, node contents={}, minimum size=0.6cm, path picture={\draw[double=orange,white,thick,double distance=1pt,shorten >=0pt] plot[variable=\t,domain=-1:1,samples=51] ({\t},{0.2*exp(-100*(\t-0.05*(#1-1))^2 - 3*\t*#2))});}
      },
    weight/.style 2 args={
        edge={#1}{#2}, node contents={\pgfmathparse{0.35*#1-#2*0.15}\pgfmathprintnumber[fixed]{\pgfmathresult}}, fill=white, inner sep=2pt
      }
  ]

  \begin{scope}[xshift=8cm]
    \nn{bayes}
  \end{scope}

  % Draw distros for all Bayesian edges.
  \foreach \i in {1,2,3}
  \foreach \j in {1,...,5}
  \path (X\i-bayes) -- (H\j-bayes) node[distro={\i}{\j}];
  \foreach \i in {1,...,5}
  \path (H\i-bayes) -- (Y-bayes) node[distro={\i}{1}];
\end{tikzpicture}

\caption{The diagram describes the structure of the Bayesian neural network used in the experiments. The developed network takes the successive values of the calculated FFT signal as input arguments, $X1$,...,$X5$, which are passed to five hidden layers, $H1$,...,$H5$, whose activation function is $\tanh$. The output of the network, $Y$, uses the sigmoid function as an activation function. In contrast to traditional neural networks, the weights are represented by distributions}
    \label{fig:bayesian_schema}
\end{figure}

For the experiment, BNNs structure was developed, as shown in Fig. \ref{fig:bayesian_schema}. In contrast to traditional neural networks, the weights are represented by distributions. The BNN that was created uses consecutive calculated FFT signal values as input arguments $X1$,...,$X5$, which are passed to five hidden layers, $H1$,...,$H5$. The activation function of these hidden layers is $\tanh$. Prior distribution for weight coefficients for all layers was selected as standard normal distribution. It is justified because of regularization properties of the normal distribution and that signals were reasonably scaled. The output $Y$ of the neural network employs the sigmoid function as its activation function \cite{Kucukelbir2016,mac2003}.

\begin{figure}[!h]
\centering
  \includegraphics[width=0.8\columnwidth]{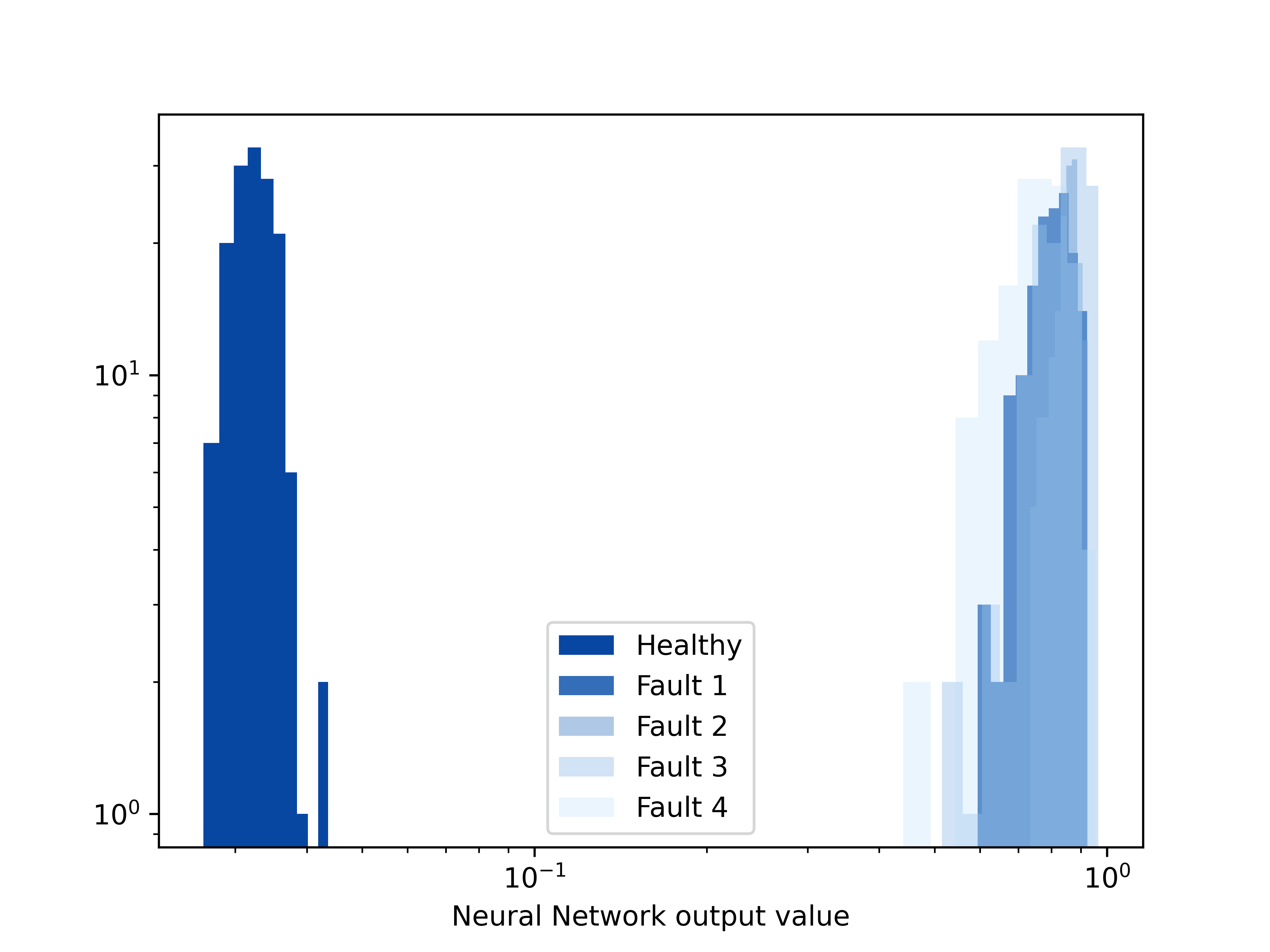}
\caption{Comparison of the distributions of output values of the designed BNN for all signal classes.
\label{fig:result_hist}}
\end{figure}

\begin{figure}[!h]
\centering
  \includegraphics[width=0.6\columnwidth]{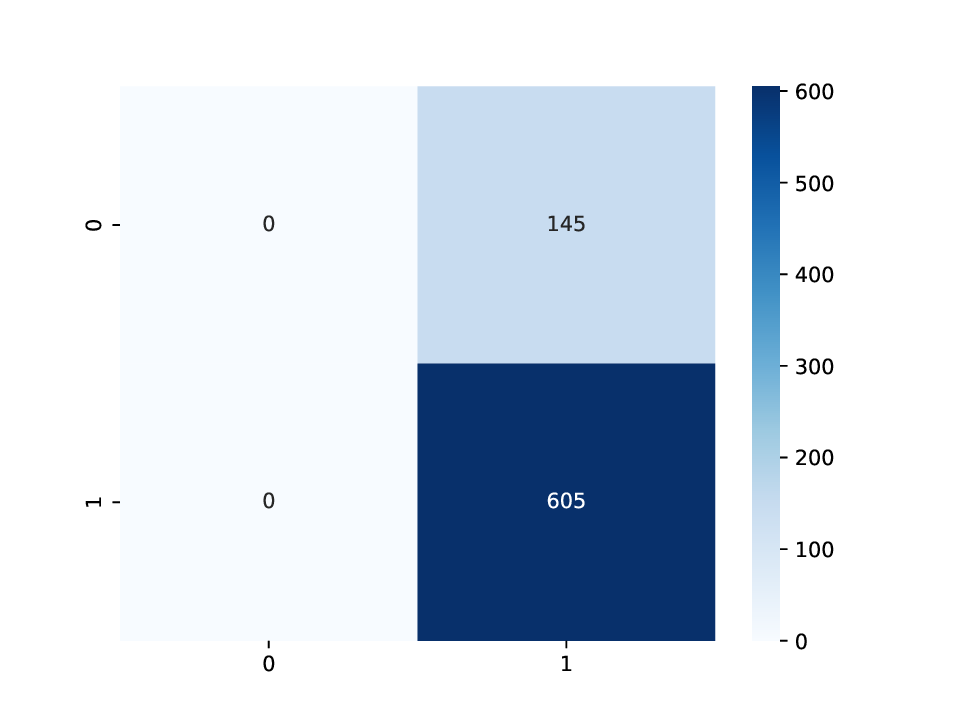}
\caption{Confusion matrix for one of the 100 trials of the developed DNN. The proposed structure flawlessly detects damages and correctly recognizes signals from undamaged devices in almost $80\%$ of cases.
\label{fig:conf_mat_dnn}}
\end{figure}

\begin{figure}[!h]
\centering
  \includegraphics[width=0.6\columnwidth]{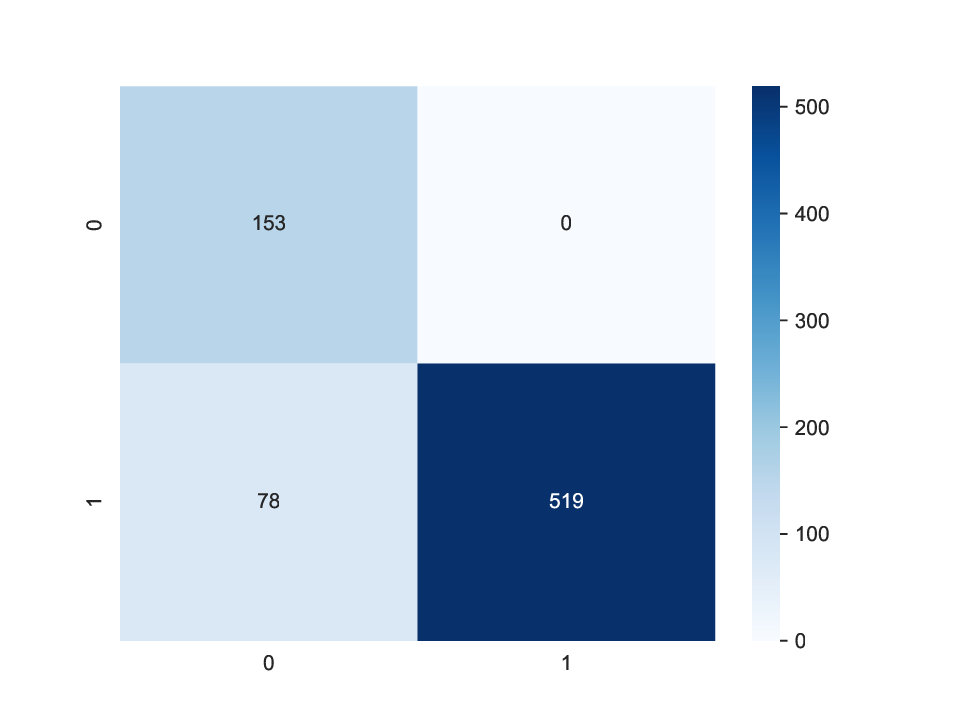}
\caption{Confusion matrix for one of the 100 trials of the developed BNN. The proposed structure flawlessly detects damages and correctly recognizes signals from undamaged devices in almost $89\%$ of cases. .
\label{fig:conf_mat}}
\end{figure}

\subsection{Experimental result}

% \begin{table}[!h]
% \centering
% \caption{Statistical distribution of predictor values for specific signal classes}
% \label{tab:com_class}
% \begin{tabular}{@{}|c|cc|@{}}
% \toprule
% Class name & Predictor mean value & Predictor standard deviation \\ \midrule
% \hline
% Healthy    & 0.046                & 0.211                        \\
% Fault 1    & 0.824                & 0.367                        \\
% Fault 2    & 0.835                & 0.368                        \\
% Fault 3    & 0.866                & 0.324                        \\
% Fault 4    & 0.779                & 0.379                        \\ \bottomrule
% \end{tabular}%

% \end{table}
% {\color{red} add pmcy description}

Due to the impossibility of implementing both types of networks using one framework, it was decided to use the pytourch library for RRN-type networks.Due to the impossibility of implementing both types of networks using one framework, it was decided to use the pytourch library for RRN-type networks. For BNN  implementation and experimentation were carried out using the PyMC 5 library \cite{pymc2023}, which enables construction of Bayesian models with a Python API and fitting these models using Markov Chain Monte Carlo (MCMC) techniques.

Each signal was assigned a class: Signals from a non-damaged device belong to class 0, while the remaining signals belong to class 1. To verify the accuracy of the developed Bayesian network, the dataset was randomly divided into a training set and a test set in the proportion 80/20.

An analysis of the results was then conducted. This process was repeated 100 times, and the average accuracy of the network was approximately $85\% $ for BNN and $80\%$ for RNN . An example confusion matrix for BNN is presented in Fig. \ref{fig:conf_mat} and for RNN in Fig. \ref{fig:conf_mat_dnn}. The displayed tables demonstrate the precision of the developed BNN. The suggested design detects damages flawlessly (top right) and accurately identifies signals from undamaged devices in nearly $89\%$ of instances. This is a desirable characteristic as it informs about malfunctions in the system with high accuracy. Looking at the confusion matrix RNN on \ref{fig:conf_mat_dnn} we can also see that the good result indicated by DNN is incorrect, due to the poorly balanced training set, in which the class of damaged devices dominates. This is typical for this type of sets, however, we see here the advantage of the BNN solution, in which we do not need to balance the training set.

\begin{figure*}[!ht]
     \centering
     \begin{subfigure}{0.4\textwidth}
         \centering
  \includegraphics[width=1\columnwidth]{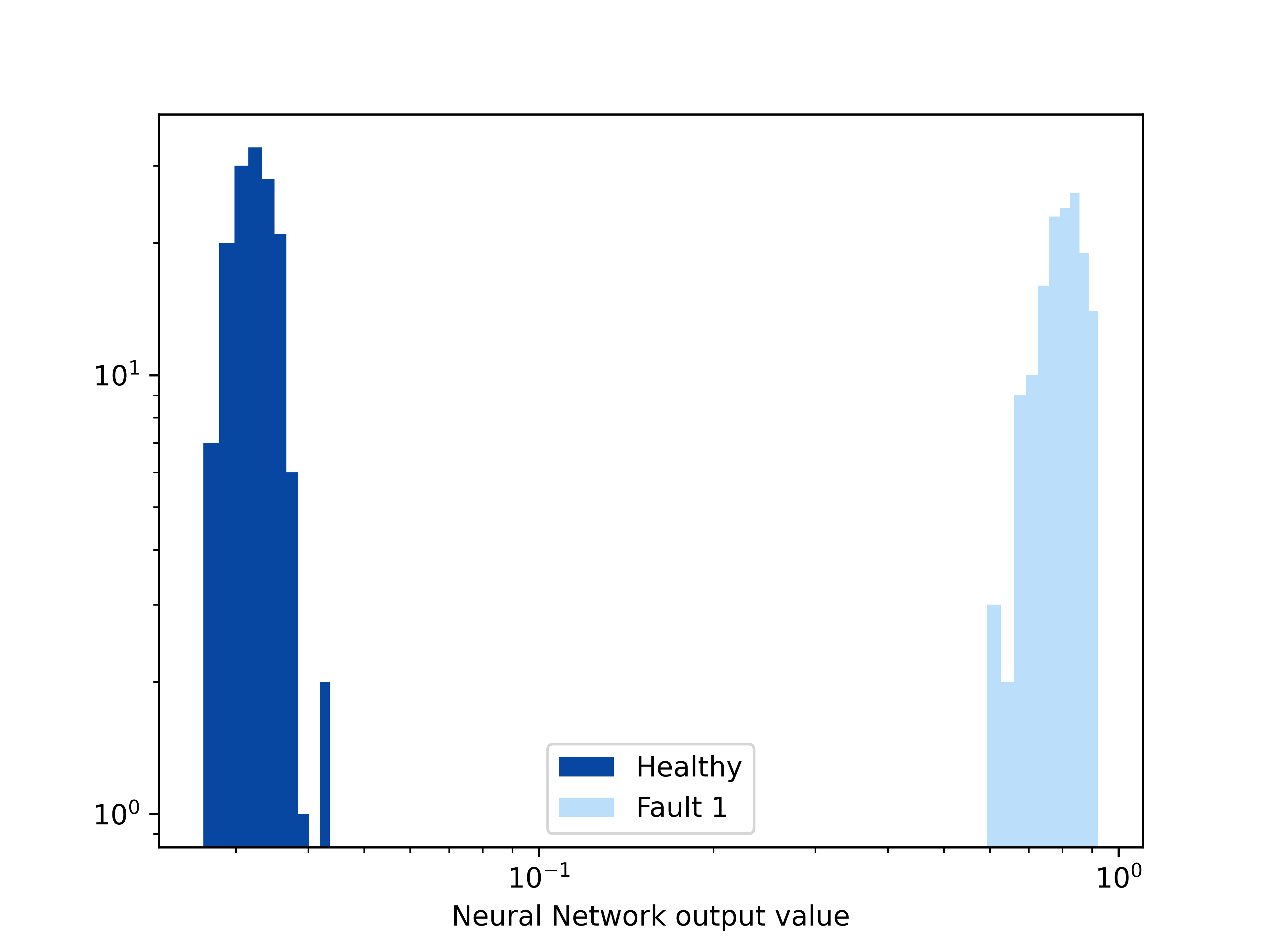}
\caption{Comparison of the distributions of output values of the designed BNN for health and fault 1 signal.\label{fig:result_hist_f1}}
\includegraphics[width=1\columnwidth]{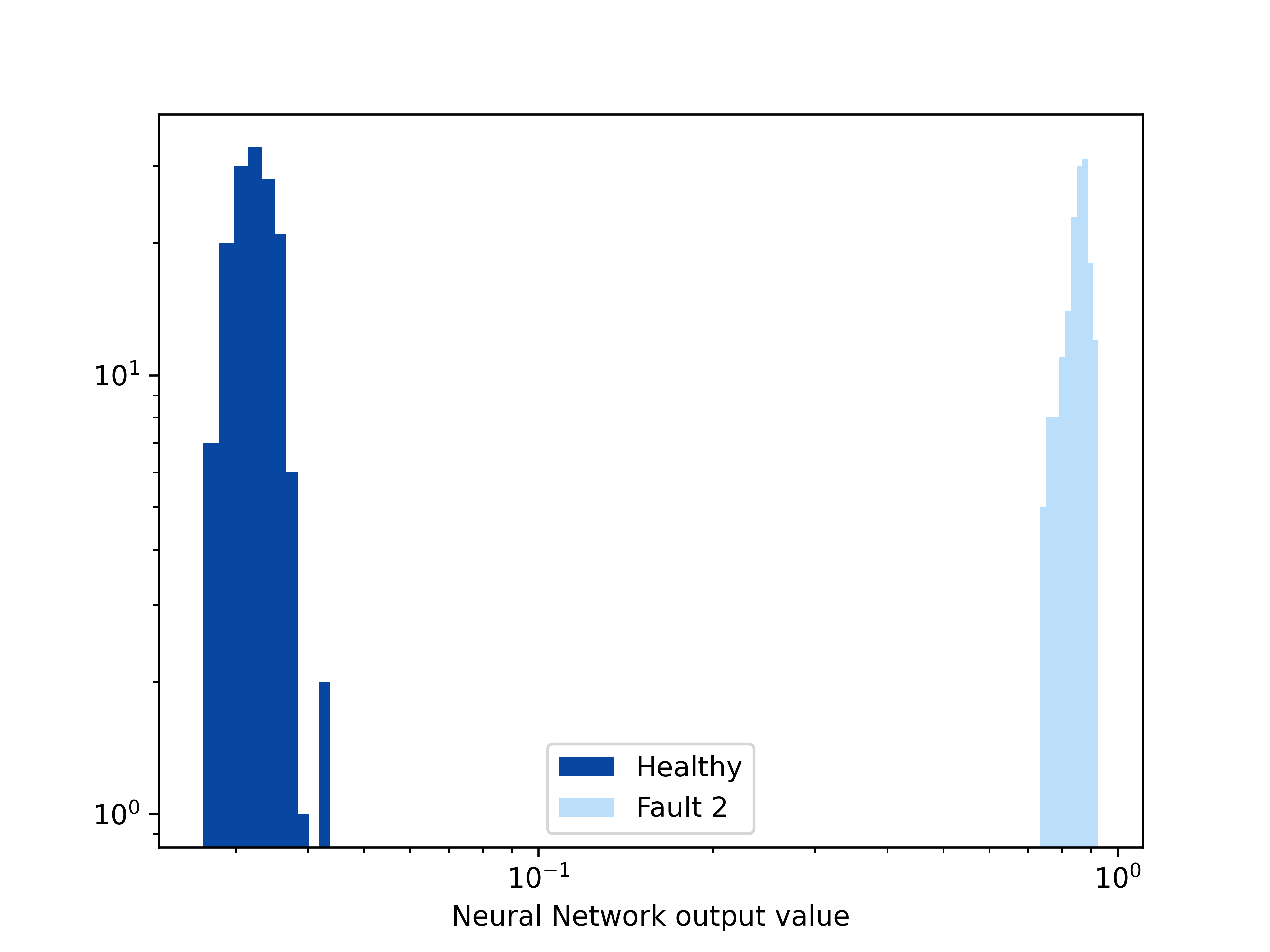}
\caption{Comparison of the distributions of output values of the designed BNN for health and fault 2 signal.\label{fig:result_hist_f2}}
     \end{subfigure}
     \hfill
     \begin{subfigure}{0.4\textwidth}
         \centering
  \includegraphics[width=1\columnwidth]{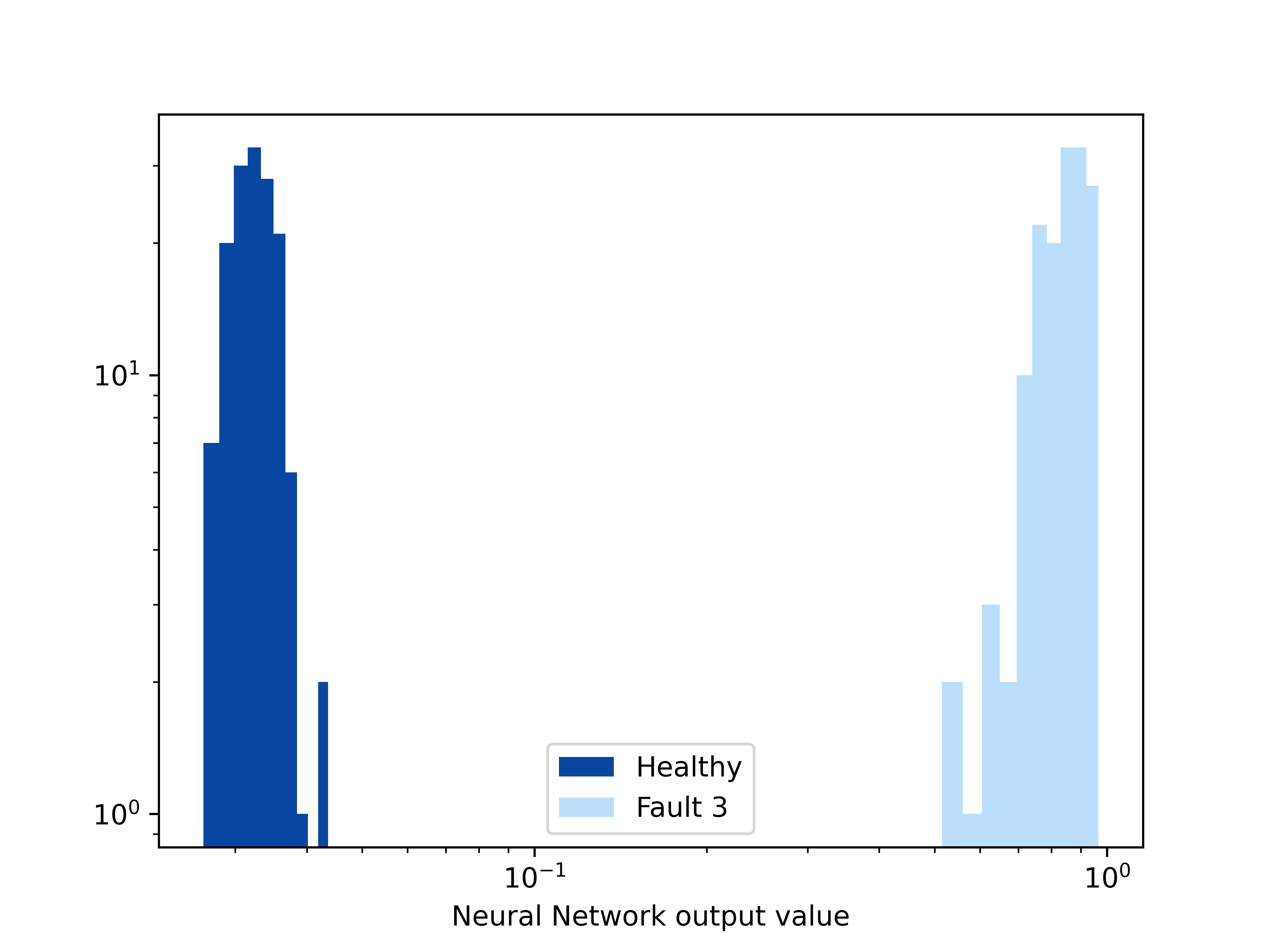}
\caption{Comparison of the distributions of output values of the designed BNN for health and fault 3 signal. \label{fig:result_hist_f3}}
  \includegraphics[width=1\columnwidth]{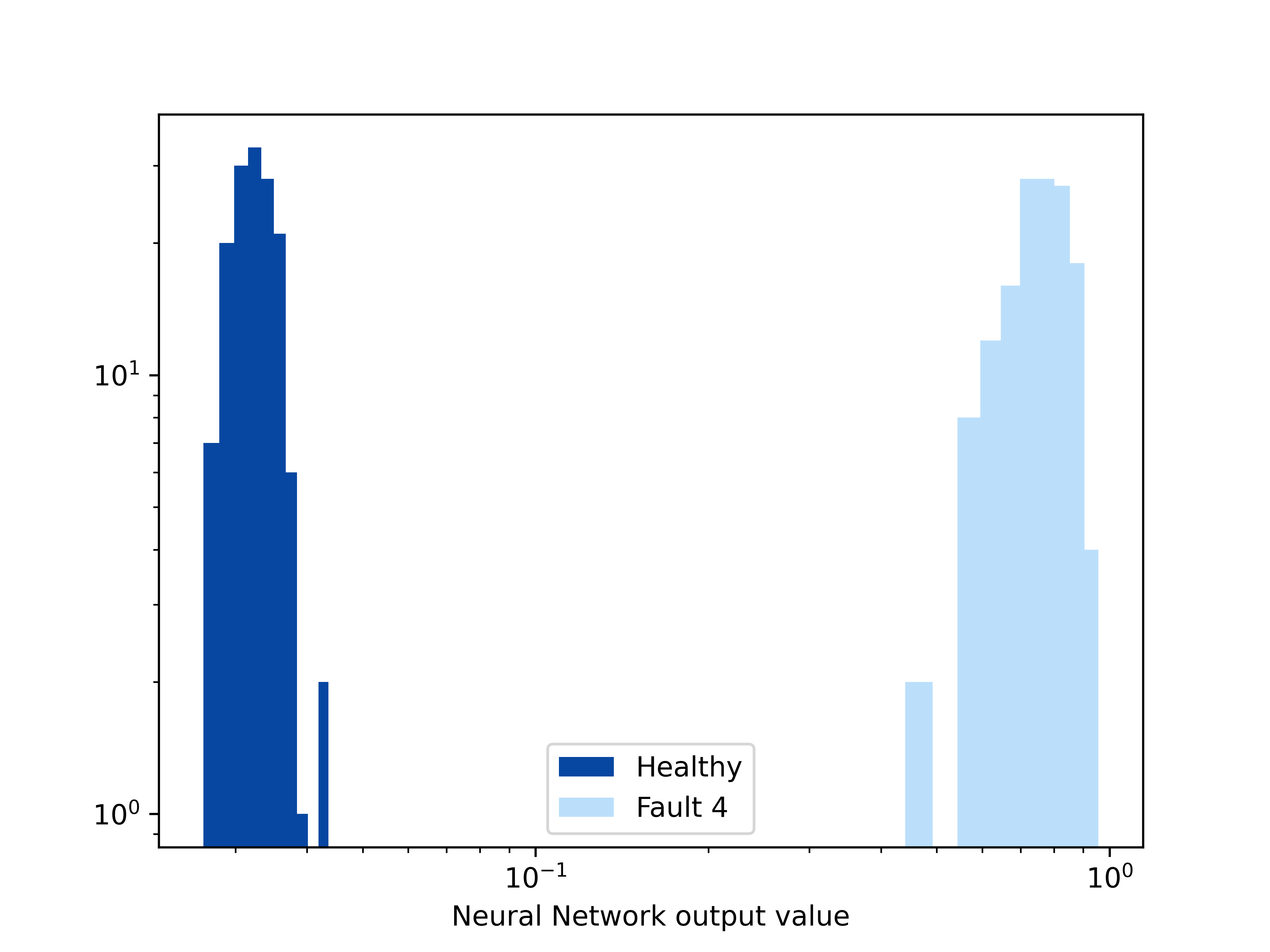}
\caption{Comparison of the distributions of output values of the designed BNN for health and fault 4 signal. \label{fig:result_hist_f4}}
     \end{subfigure}
     \hfill
        \caption{Comparison of the distributions of output values of the designed BNN for a healthy signal and individual classes of faults.}
        \label{fig:com_hist}
\end{figure*}

Additionally, as shown in Fig. \ref{fig:com_hist}, we can see that in BNN we can interpret statistical distributions of features, which is not possible in the DNN architecture.

\section{Conclusion}
\label{sec:Conclusions}
Bayesian neural networks (BNNs) and Deep neural networks (DNN) has been applied to fault detection. BNNs detect faults more accurately than traditional neural networks, even for imbalanced or incomplete datasets. In this paper, we focus on fault diagnosis of electric motors using acoustic signals. Such diagnostic signals in time-domain have varying length and are sampled with high frequency, thus necessitating a new signal representation for classification. 

We compared the classification of electric motor faults for small household appliances using Bayesian neural network and deep neural network. BNN developed for fault classification proved to be $100\%$ accurate in detecting faulty devices and $70\%$ accurate in detecting healthy devices. DNN proved to be effective - proved to be $100\%$ accurate in detecting faulty devices but was unable to diagnose a properly working device. Additionally, the results obtained by Bayesian neural network are better interpreted due to the statistical nature of this solution. This allows us to conclude that this type of network can be potentially useful in case of diagnostic problems and is suitable for developing interpretable systems for industrial diagnostic systems.

The open problem for analysis is the comparison of different architectures with respect to the resources needed for their training and classification. The next step would be to develop a set of benchmarks and metrics that would help in comparing different architectures. In addition, there are plans to attempt to develop classifiers capable of distinguishing between different classes of faults for different neural network architectures and compare their performance.

% \appendix
% \section{Example Appendix Section}


%%       \cite{<label>} ==> [1]

%%
% Example citation, See \cite{lamport94}.


\section*{Acknowledgement}
Work partially realised in the scope of project titled ''Process Fault Prediction and Detection''. Project was financed by The National Science Centre on the base of decision no. UMO-2021/41/B/ST7/03851. Part of work was funded by AGH’s Research University Excellence Initiative under projects: Waldemar Bauer PAKIET HABILITACYJNY -edycja II  and “DUDU - Diagnostyka Uszkodzeń i Degradacji Urządzeń''.

\bibliographystyle{elsarticle-num} 
\bibliography{bibliography}

\end{document}